\documentclass[11pt]{elsarticle}

\usepackage{lineno,hyperref}

\usepackage{float}
\usepackage{graphicx}
\usepackage{subfig}
\usepackage{mathrsfs}
\usepackage{graphics}
\usepackage{float}
\usepackage{amsmath, commath}
\usepackage{amsfonts}
\usepackage{mathrsfs}
\usepackage{amsmath}
\usepackage[table]{xcolor}
\usepackage{longtable}
\usepackage{optidef}

\usepackage{optidef}
\usepackage{graphicx}
\usepackage{float}
\usepackage{graphicx}
\usepackage{subfig}
\usepackage{mathrsfs}
\usepackage{float}
\usepackage{amsmath}
\usepackage{amsfonts}
\usepackage{booktabs}
\usepackage{natbib}
\usepackage{bbm}
\usepackage{algorithm}
\usepackage[noend]{algpseudocode}
\usepackage{multirow}
\algdef{SE}[DOWHILE]{Do}{doWhile}{\algorithmicdo}[1]{\algorithmicwhile\ #1}

\usepackage[letterpaper, portrait, margin=1in]{geometry}

\usepackage{booktabs}
\usepackage{tabularx}

\usepackage{xcolor}
\hypersetup{
    colorlinks,
    linkcolor={red!50!red},
    citecolor={blue!50!blue},
    urlcolor={blue!80!blue}
}

\usepackage{optidef}

\usepackage{amsthm}

\theoremstyle{definition}

\makeatletter

\usepackage{newtxtext} 

\DeclareMathOperator*{\argmax}{arg\,max}

\makeatletter
\newcommand*{\centerfloat}{%
  \parindent \z@
  \leftskip \z@ \@plus 1fil \@minus \marginparwidth
  \rightskip \leftskip
  \parfillskip \z@skip}
\makeatother

\usepackage{amssymb,amsmath,caption}
\usepackage[hang,flushmargin]{footmisc}
\newcommand{\algorithmfootnote}[2][\footnotesize]{
  \let\old@algocf@finish\@algocf@finish
  \def\@algocf@finish{\old@algocf@finish
    \leavevmode\rlap{\begin{minipage}{\linewidth}
    #1#2
    \end{minipage}}%
  }%
}

\usepackage{setspace}
\usepackage{natbib}
\usepackage{algorithm}
\usepackage[noend]{algpseudocode}
\algdef{SE}[DOWHILE]{Do}{doWhile}{\algorithmicdo}[1]{\algorithmicwhile\ #1}

\usepackage{appendix}
\usepackage{array}
\usepackage{threeparttable}
\usepackage{multirow}
\usepackage{rotating}

\usepackage{makecell}

\journal{}

\biboptions{authoryear}

\begin{document}
\begin{frontmatter}
\title{Large Language Models for Travel Behavior Prediction}

\author[label0,label1]{Baichuan Mo}
\author[label2]{Hanyong Xu\corref{mycorrespondingauthor}}
\author[label3]{Ruoyun Ma}
\author[label1]{Jung-Hoon Cho}
\author[label1]{Dingyi Zhuang}
\author[label1]{Xiaotong Guo}
\author[label2]{Jinhua Zhao}
\address[label0]{Department of Civil Engineering, Tsinghua University, Beijing, China, 100084}
\address[label1]{Department of Civil and Environmental Engineering, Massachusetts Institute of Technology, Cambridge, MA 02139}
\address[label2]{Department of Urban Studies and Planning, Massachusetts Institute of Technology, Cambridge, MA 20139}
\address[label3]{Department of Management Science and Engineering, Stanford University, Stanford, CA 94305}
\cortext[mycorrespondingauthor]{Corresponding author}

\begin{abstract}
Travel behavior prediction is a core problem in transportation demand management and is traditionally addressed using numerical models calibrated on observed data. With recent advances in large language models (LLMs), new opportunities have emerged to model human decision-making through natural language reasoning. This study explores the use of LLMs for travel behavior prediction through two complementary frameworks. The first framework employs a zero-shot prompting strategy, where the prediction task, traveler attributes, and relevant domain knowledge are described in text, enabling the LLM to directly generate predictions without task-specific training data. The second framework uses LLM-generated text embeddings as high-level representations of travel scenarios, which are then combined with conventional supervised learning models to support prediction in small-sample settings. Empirical results show that both approaches achieve performance comparable to, and in some cases competitive with, classical models such as multinomial logit, random forest, and neural networks. These findings suggest that LLMs offer a flexible and data-efficient alternative for travel behavior prediction.
\end{abstract}

\begin{keyword}
Large language model; Travel behavior prediction 
\end{keyword}

\end{frontmatter}


\section{Introduction}
Understanding human travel behavior has always been a crucial part of transportation planning. Travel behavior is a broad concept that consists of many attributes, such as the mode of transportation, the purpose of the travel, the choice of destination, and the time of departure. Consider travel mode choice as an example, which refers to how people make decisions on their means of travel. In the task of predicting mode choices, making accurate predictions is useful for both agencies and individuals. For agencies, it facilitates providing better transit services and forecasting travel conditions in the short term, while informing transportation infrastructure investments, better policies, and land use planning in the long run. In the traditional Four-Step Model for future demand and performance analysis, the mode share of a given population is estimated using mode split models \citep{Ben_Akiva_dc1985}. For individuals, on the other hand, mode choice prediction embedded in map applications helps them to make sound travel decisions. Conventional mode choice analysis relies on using numerical data to construct a mathematical model to represent human preferences \citep{mo2018impact, mo2021impacts}. One example is the discrete choice model, where mode choice probabilities are calculated assuming people make choices to maximize their utilities \citep{Ben_Akiva_dc1985}.

Recently, large language models (LLMs) have generated a tremendous amount of excitement in research due to their advanced language interaction and interpretation capabilities exhibited by-products such as ChatGPT. Having been trained on extremely large corpus and with parameters to the size of more human hundreds of billions, LLMs are able to interpret and execute diverse types of intricate tasks with a certain level of intelligence \citep{Zhao_Zhou_2023}. In transportation, studies have shown the abilities of GPT in making mobility predictions which outperforms state-of-the-art time-series machine learning models \citep{Wang_Fang_Zeng_Cheng_2023}. However, the question of whether LLMs can tackle the problem of travel behavior predictions is unclear. 

In response to the above inquiry, this study investigates the capability of large language models (LLMs) to predict individual travel behavior, with travel mode choice and trip purpose prediction serving as a representative case studies. We propose two distinct LLM-based prediction frameworks. The first framework adopts a zero-shot prompting approach, in which carefully designed prompts describe the travel mode choice task, individual travel characteristics, and traveler attributes, while explicitly incorporating transportation domain knowledge to guide the model’s reasoning. This framework enables direct prediction through text generation without requiring any task-specific training data. The second framework leverages LLM-generated text embeddings as high-level feature representations, which are subsequently used as inputs to supervised learning models for prediction under small-sample settings. Empirical results demonstrate that both frameworks can achieve performance comparable to, and in some cases competitive with, classical travel behavior models such as multinomial logit, random forest, and neural networks, highlighting the potential of LLMs as flexible and data-efficient tools for travel behavior prediction.


The remainder of this paper is organized as follows. The literature review is presented in Section \ref{sec_liter}. In Section \ref{sec_method}, we describe the method for using LLM to predict travel behavior. Case studies and experiment results are shown in Section \ref{sec_case}. Section \ref{sec_conclusion} concludes the paper and discusses the future research directions.

\section{Related work}\label{sec_liter}
\subsection{Recent developments in large language models}

Large Language Models (LLMs) typically refer to the language models based on the Transformer architecture with more than hundreds of billions of parameters, which have been shown to have abilities to excel in many complex tasks \citep{Zhao_Zhou_2023}. \citet{Wei_2022} has demonstrated the emerging abilities of LLMs that only appear in larger models and cannot be predicted based on smaller models' performance. Newer models such as GPT-3.5, GPT-4 \citep{OpenAI_Achiam_Adler_Agarwal_Ahmad_Akkaya_Aleman_Almeida_Altenschmidt_Altman_2024}, Gemini \citep{Team_Anil_Borgeaud_Alayrac_Yu_Soricut_Schalkwyk_Dai_Hauth_Millican_2024}, and Llama \citep{Touvron_Lavril_Izacard_Martinet_Lachaux_Lacroix_Rozière_Goyal_Hambro_Azhar_2023, Touvron_Martin_Stone_Albert_et_2023, Grattafiori_Dubey_Jauhri_Pandey_Kadian_AlDahle_Letman_Mathur_Schelten_Vaughan_2024} illustrated strong capabilities in language processing and manipulation, quantitative reasoning, as well as planning and learning \citep{Bubeck_2023}. Moreover, \citet{Gurnee_Tegmark_2023} has shown LLMs do have structured knowledge on both space and time using Llama-2 family models; they illustrated the model activations demonstrate linear representations of landmarks and historical events. In addition, products like ChatGPT which implement LLMs into chatbots and search engines that are easily accessible by the general public generated tremendous attention and excitement in this field \citep{Zhao_Zhou_2023}. 

Apart from their powerful ability in complex tasks, LLM has brought a new research paradigm that leverages the interaction with prompt interfaces or APIs to develop applications as opposed to trained models from scratch \citep{Zhao_Zhou_2023}. The performances of these LLMs to user inputs often depend on the level of prompt engineering. Common methods to improve the performance of the tasks in LLMs include in-context learning, which leverages formulations of tasks and demonstrations into natural languages \citep{Brown_Mann_2020, Dong_Li_2023}, chain-of-thought, which introduces intermediate reasoning \citep{ Chu_Chen_Chen_Yu_He_Wang_Peng_Liu_Qin_Liu_2024}, and plan-and-solve \citep{Wang_Xu_Lan_Hu_Lan_Lee_Lim_2023, Zhou_Schärli_2023} to breakdown complex tasks and elicit the step-by-step reasoning power of LLMs \citep{Zhao_Zhou_2023}. 

On the other hand, apart from the fundamental abilities of LLMs in language generation, knowledge utilization, and complex reasoning, LLMs are also prone to several drawbacks that need to be considered when utilizing them \citep{Zhao_Zhou_2023}. One limitation is Hallucination, where texts generated by LLMs could conflict with themselves or can not be verified \citep{Bang_Cahyawijaya_2023, Huang_Yu_Ma_Zhong_Feng_Wang_Chen_Peng_Feng_Qin_et_2025}. Another challenge of LLMs is their limit on recent knowledge since models trained on a big corpus of data with a large number of parameters typically cannot be updated in real-time \citep{Zhao_Zhou_2023}. A third limitation is inconsistent reasoning, where LLMs' reasoning and answer are not always aligned \citep{wei2022chain}.

\subsection{Large language model as a predictor in human mobility}
With the exceptional performance of LLMs, studies in many fields have utilized them to liberate traditional practices. ChatGPT has been shown to achieve high accuracies for linguistic-related classification tasks, such as genre recognition, personality prediction, political opinions, and hateful speech detection \citep{Liu_Han_Ma_2023, Kuzman_Mozetič_Ljubešić_2023,  Amin_Cambria_Schuller_2023, Zhang_Ding_Jing_2023, Huang_Kwak_An_2023}. 
Similarly, the utilization of LLMs in transportation also boomed in the past several year, including tasks such as forecasting in traffic, human mobility, and demand, as well as data imputation \citep{Zhang_Sun_Wang_Nie_Ma_Li_Sun_Ban_2025}. 

Specifically, in the realm of human mobility, a series of experiments have demonstrated the semantic and reasoning power of LLMs.
Before the rise of GPT family, several studies used pre-trained language models to predict mobility. \citet{Xue_Voutharoja_Salim_2022, Xue_Salim_2022} have been looking at converting mobility data into language descriptions to fine-tune language models, while \cite{Kobayashi_Takeda_Yamazaki_Kamisaka_2023}, \cite{Gong_Lin_Zhang_Lu_Han_Liu_Guo_Lin_Wan}, and \cite{Wu_He_Wang_Wang_2024} examined encodings and embeddings in mobility prediction in the transformer architecture. With available commercial LLMs, \citet{Luo_Cao_Jin_Liu_Yin_2024} proved GPT's ability to infer trajectory patterns from data, while \citet{Wang_Fang_Zeng_Cheng_2023} initially demonstrated the powerful ability of GPTs to infer the next location with the understanding of semantic sequential mobility data and prompts converted from historical locations. The predictions by LLMs outperform the traditional time-series models such as LSTM and the Multi-Head Self-Attentional (MHSA) neural network. Additional studies illustrated how incorporating behavioral theories \citep{Shao_Xu_Fan_Ding_Yuan_Wang_Li_2024}, agentic logical thinking \citep{Wang_Jiang_Yang_Wu_Onizuka_Shibasaki_Koshizuka_Xiao_2024, Li_Huang_Lv_Xiao_Li_Yue_2024, Li_Feng_Chi_Hu_Zhao_Xu_2024, Feng_Du_Zhao_Li_2025}, pattern seeking \citep{Qin_Zhang_Wang_Ma_2025, Wang_Jiang_Yang_Wu_Onizuka_Shibasaki_Koshizuka_Xiao_2024, Li_Huang_Lv_Xiao_Li_Yue_2024}, and understanding of additional semantic data \cite{Liang_Liu_Wang_Zhao_2023} also facilitates the trajectory or time-series predictions.

LLMs' planning ability has been popular in solving trip planning, one of the downstream applications of transport prediction. Leveraging multiple sources and queries \citep{Xie_Zhang_Chen_Zhu_Lou_Tian_Xiao_Su_2024, Fang_Yang_Wang_Di_2024, Singh_Verma_Wang_Bharadwaj_Fashandi_Ferreira_Lee_2024} and breaking down into multiple steps \citep{Tang_Wang_Qu_Yan_Wu_Zhuang_Kai_Hou_Guo_Zheng_2024,Xie_Zou_2024} have been shown relatively successful.

Despite existing research offers promising potential for using LLMs as predictors, the existing literature only focuses on leveraging LLMs as a predictor in sequential mobility prediction and there has been very few research exploring the ability of LLMs in travel behavior prediction tasks. In terms of travel choice prediction, a few studies analyzed how to incorporate historical personalized information to improve prediction \citep{Zhai_Tian_Li_Zhao_2024, Wang_Duan_He_Lyu_Chen_Zheng_Yao_Ma_2024, Chen_He_Wang_Chen_Luo_2024}. \cite{Liu_Li_Yin_2024} introduced persona loading and few-shots to improve the misalignment of LLM predictions and the human behavior. However, there hasn't been a systematic study on understanding the zero-shot learning in travel behavior predictions, especially in terms of the tasks of mode choice and travel purpose inference with LLM, which we will focus in this study.


\section{Methodology}\label{sec_method}
\subsection{Preliminaries}
Large language models (LLMs) are neural network-based architectures designed to process and generate human language text. Recent LLMs are usually built upon the transformer architecture. This architecture employs a deep neural network with self-attention mechanisms, allowing for the modeling of long-range dependencies in sequences of text. The size and scale of LLMs have grown exponentially, with models containing hundreds of millions or even billions of parameters. The increase in model size directly correlates with improved performance on various NLP tasks.

Pre-training is a crucial phase in the development of LLMs, where the model is exposed to massive amounts of text data to learn the statistical properties of language, including grammar, semantics, and world knowledge. The primary objective of pre-training is to train the model to predict the next word in a sentence or sequence of words. This task, known as language modeling, allows the model to capture the statistical regularities and contextual dependencies within the training data. Note that different LLMs (e.g., BERT and GPT) may have different pre-training tasks. In this study, we introduce the pre-training based on GPT's framework for unsupervised multitask learning \citep{radford2019language}. The objective of pre-training can be expressed as:
\begin{align}
\boldsymbol{\theta^*} = \arg\max_{\boldsymbol{\theta}} \sum_{u\in\mathcal{U}}\sum_{i=1}^{n} \log \mathbb{P}\left(w_i^{(u)} \mid w_1^{(u)}, w_2^{(u)}, \ldots, w_{i-1}^{(u)}; \boldsymbol{\theta}\right)
\end{align}
where $\mathcal{U}$ is the set of all training corpus. $w_i^{(u)}$ is the $i$-th token of the $u$-th sequence. Given the trained parameter $\boldsymbol{\theta^*}$, we can use the model to generate answers for various tasks:
\begin{align}
   \boldsymbol{s}^* = \argmax_{\boldsymbol{s}} \mathbb{P}(\boldsymbol{s}\mid\text{(Input, Task)};\boldsymbol{\theta^*}) \label{eq_example1}
\end{align}
where $\boldsymbol{s}^*$ is the output sequence with the largest probability (or relatively large probability depending on the searching algorithm and degree of randomness). $\boldsymbol{s}^*$ is generated word by word until the ``[End]'' token is found.
For example, asking the LLM to solve an addition problem can be expressed as 
\begin{align}
   \text{``The answer is 28 [End]''} = \argmax_{\boldsymbol{s}} \mathbb{P}(\boldsymbol{s}\mid\text{``What is 3+25? [End]''};\boldsymbol{\theta^*})
\end{align}
In the real-world implementation, the ``[End]'' token will be automatically added to the input sequence and will not be displayed at the end of the output sequence. 

Besides generating text sequences, LLMs can also be used to obtain text embeddings using the internal layers before the final text output layer. Specifically, denote the LLM's with embedding outputs as $f(\cdot;\boldsymbol{\theta}^*)$. Then the text embedding (denoted as $\boldsymbol{h}$) for the input task can be obtained as:
\begin{align}
  \boldsymbol{h} = f\left(\text{(Input, Task)};\boldsymbol{\theta}^*\right)
\end{align}
$\boldsymbol{h}\in\mathbb{R}^{H}$ is usually use as features for many downstream supervised learning tasks as the extracted information of text. Its dimension $H$ depends on specific LLM architectures. Given the generalizability of LLMs, $\boldsymbol{h}$ could potentially increase the prediction performance of models with small sample sizes.   

\subsection{Conceptual framework}
Given LLMs provide a generalized multitask solver, it is possible to use them as a predictor for people's travel behavior prediction. In this study, we propose two different frameworks to incorporate LLM.

The ``(Input, Task)'' in Eq. \ref{eq_example1} is referred to as \textbf{prompt} in LLMs. The performance of LLMs for many tasks highly depends on the design of the prompt. We carefully design our prompts that include 1) task descriptions, 2) structural data, 3) descriptive data, and 4) guides for thinking with domain knowledge. Details will be shown in Section \ref{sec_prompt}.

\begin{figure}[H]
    \centering
    \includegraphics[width=1.0\linewidth]{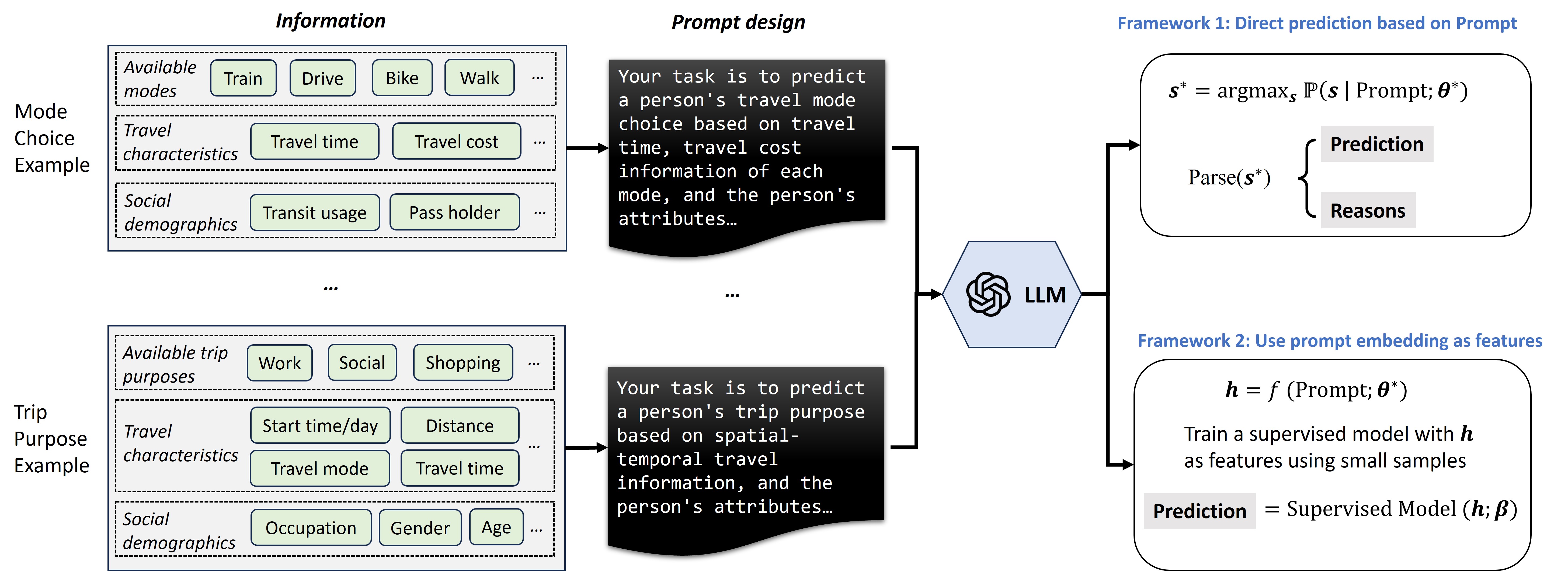}
    \caption{Conceptual framework}
    \label{fig_framework}
\end{figure}

The frameworks of the LLM-based travel behavior prediction is shown in Figure \ref{fig_framework}. Here, we use a travel mode choice task and a travel purpose prediction task as the illustration examples. The input information will be organized and embedded into the prompt (i.e., prompt design). The prompt is fed into an LLM and outputs the prediction and associated reasons. It is worth noting that no supervised training step is needed. 

After processing of LLM, we may get two different outputs depending on which prediction framework we use. For direct prediction based on prompts, we will use text-generation LLM (e.g., GPT 3.5) and ask the model to generate the predicted results and reasons. In terms of using prompt embedding as features, we will get a hidden vector $\boldsymbol{h}_n$ from the LLM (e.g., text-embedding model \citep{openai_text_embedding_3_small} in the Open AI platform) for sample $n$. Then we train a supervised learning model (e.g., multinomial logit) using $\boldsymbol{h}_n$ as features, and use the trained model to predict the travel behavior.

\subsection{Prompt design}\label{sec_prompt}
Building upon existing prompting strategies, we carefully develop context-inclusive prompts that incorporate relevant contextual information to enhance travel behavior prediction. Detailed components are described as follows.

\subsubsection{Task description}
\begin{figure}[H]
    \centering
    \includegraphics[width=0.9\linewidth]{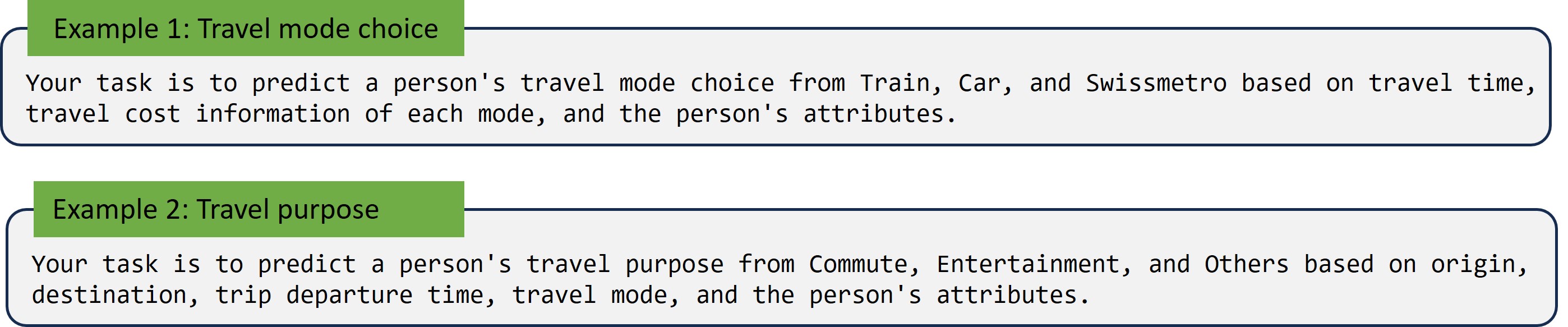}
    \caption{Example prompts for task descriptions}
    \label{fig_example_task}
\end{figure}

The task in this study is travel behavior prediction, which may include travel mode choice, travel purposes, travel departure time, duration, destination, etc. In the task description, we need to specify the available options and potential information we will provide. Example prompts are shown in Figure \ref{fig_example_task}

\subsubsection{Structural data}
\begin{figure}[H]
    \centering
    \includegraphics[width=0.9\linewidth]{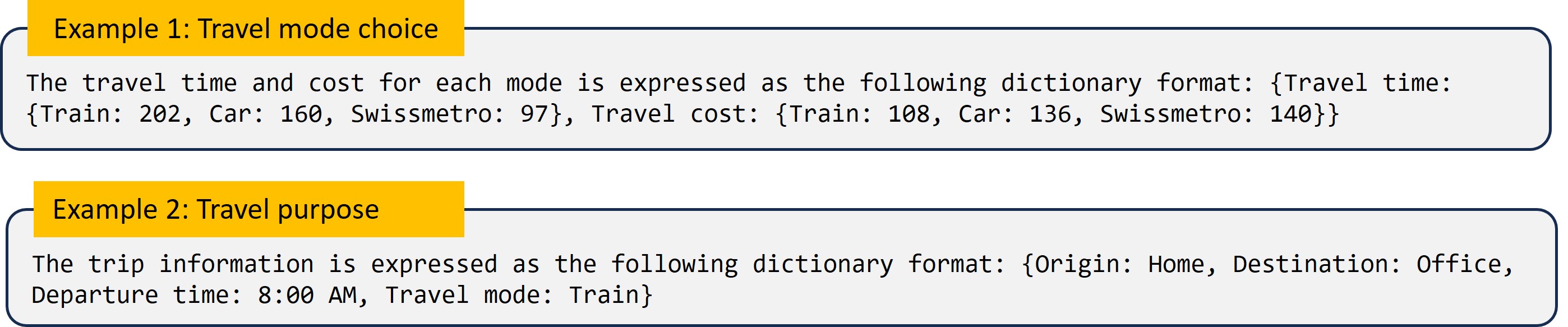}
    \caption{Example prompts for including travel characteristics}
    \label{fig_example_travel_cha}
\end{figure}

People's travel behavior is largely decided by travel characteristics, such as origin, destination, travel time, travel cost, etc. The prompts should include this information in a concise and organized way. In this study, we use a dictionary format the organize the travel characteristics information. Examples are shown in Figure \ref{fig_example_travel_cha}. 

\subsubsection{Descriptive data}
\begin{figure}[H]
    \centering
    \includegraphics[width=0.9\linewidth]{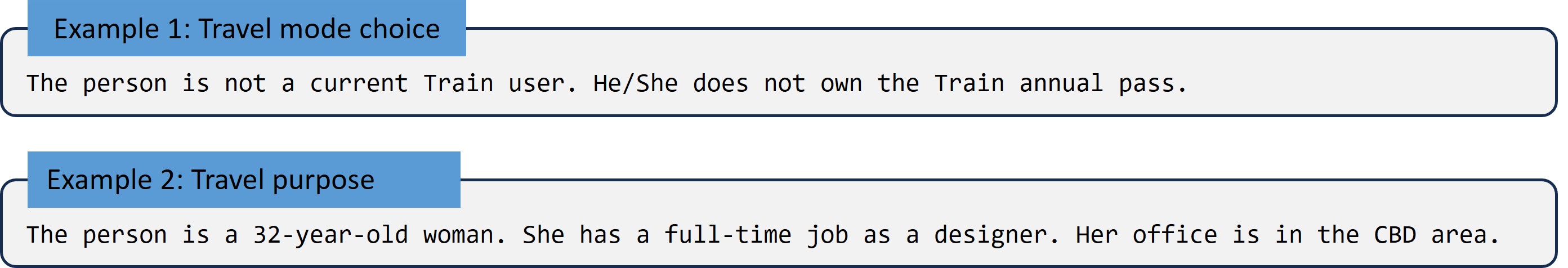}
    \caption{Example prompts for individual attributes}
    \label{fig_example_ind_att}
\end{figure}

Individual attributes may also affect people's travel behavior. This information is included in a descriptive way (instead of using dummy binary variables as typical mathematical models).  Examples are shown in Figure \ref{fig_example_ind_att}. 

\subsubsection{Guide of thinking with domain knowledge}
\begin{figure}[H]
    \centering
    \includegraphics[width=0.9\linewidth]{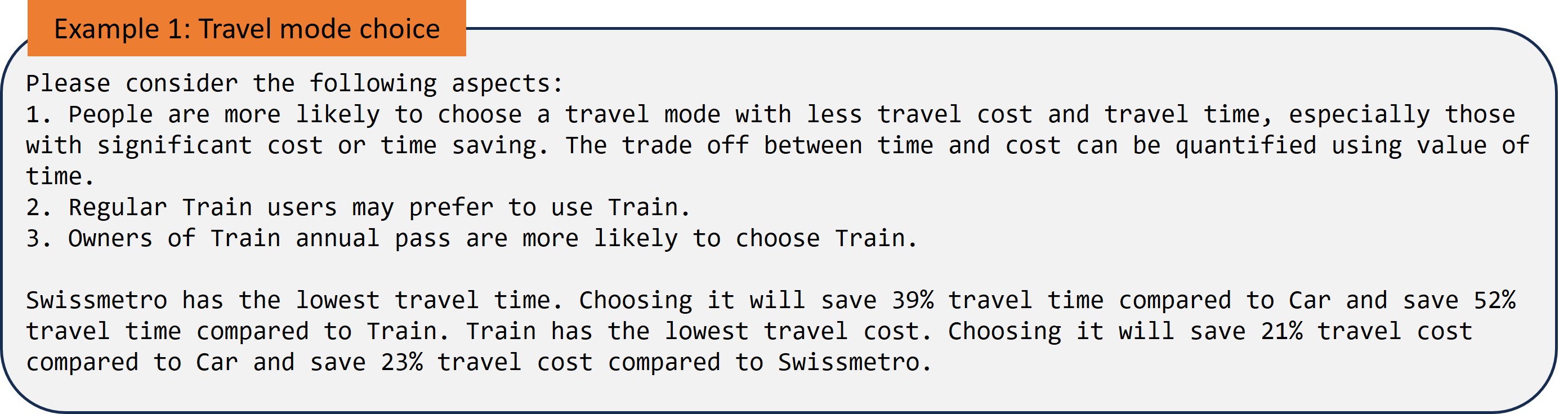}
    \caption{Example prompts for the guide of thinking with domain knowledge}
    \label{fig_example_guide_think}
\end{figure}

The ability of LLMs to perform complex reasoning can be improved by designing prompts with strategies like Chain-of-Thought \citep{Wei_2022} and Plan-to-Solve \citep{Wang_Xu_Lan_Hu_Lan_Lee_Lim_2023}. The essential idea is to guide LLMs on how to use the given information. This is the most important part of enabling LLMs to predict travel behavior without given any training data. The prompts should include domain knowledge with human common sense. The example of travel mode choice is shown in Figure \ref{fig_example_guide_think}.

The first three aspects are domain knowledge. This can be customized based on input features and the specific task. The last paragraph is used to guide LLMs for numerical comparison. It is known that arithmetic and symbolic reasoning are challenging tasks for LLMs without well-designed prompts \citep{rae2021scaling}. Even for a simple task like providing LLMs with three numbers A, B, and C, and asking them to sort the numbers, LLMs can make many mistakes (this problem may be mitigated with more advanced LLMs). Therefore, we add the last paragraph, telling LLMs which mode has the lowest travel time or travel cost, and how large the relative differences are. Experiments show that this tends to be an effective way to enable LLMs for arithmetic reasoning in this context. 

\subsubsection{Interpretation and output}
\begin{figure}[H]
    \centering
    \includegraphics[width=0.9\linewidth]{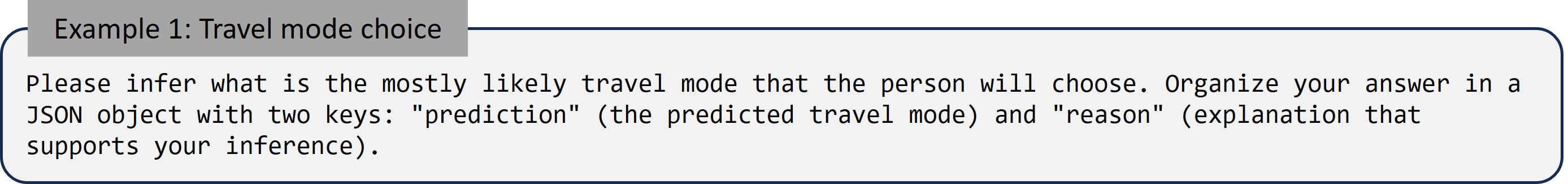}
    \caption{Example prompts for interpretation and output}
    \label{fig_example_prompt_output}
\end{figure}

The last part of the prompt is to specify the outputs of the LLMs. The outputs include two parts: 1) the predictions and 2) the reasons that support the prediction. There are two motivations to ask for the reasons. First, it gives the model interpretability and helps us to diagnose the prompt design. Moreover, asking for explanations also improves LLMs' reasoning ability. This is similar to the Plan-to-Solve prompting \citep{Wang_Xu_Lan_Hu_Lan_Lee_Lim_2023}, where the LLM's reasoning performance is improved when the specific calculation steps are asked to output. The example prompt for travel behavior prediction is shown in Figure \ref{fig_example_prompt_output}. Note that we also specify the output format so that we can easily extract the results.

\subsubsection{Summary}
\begin{figure}[H]
    \centering
    \includegraphics[width=0.9\linewidth]{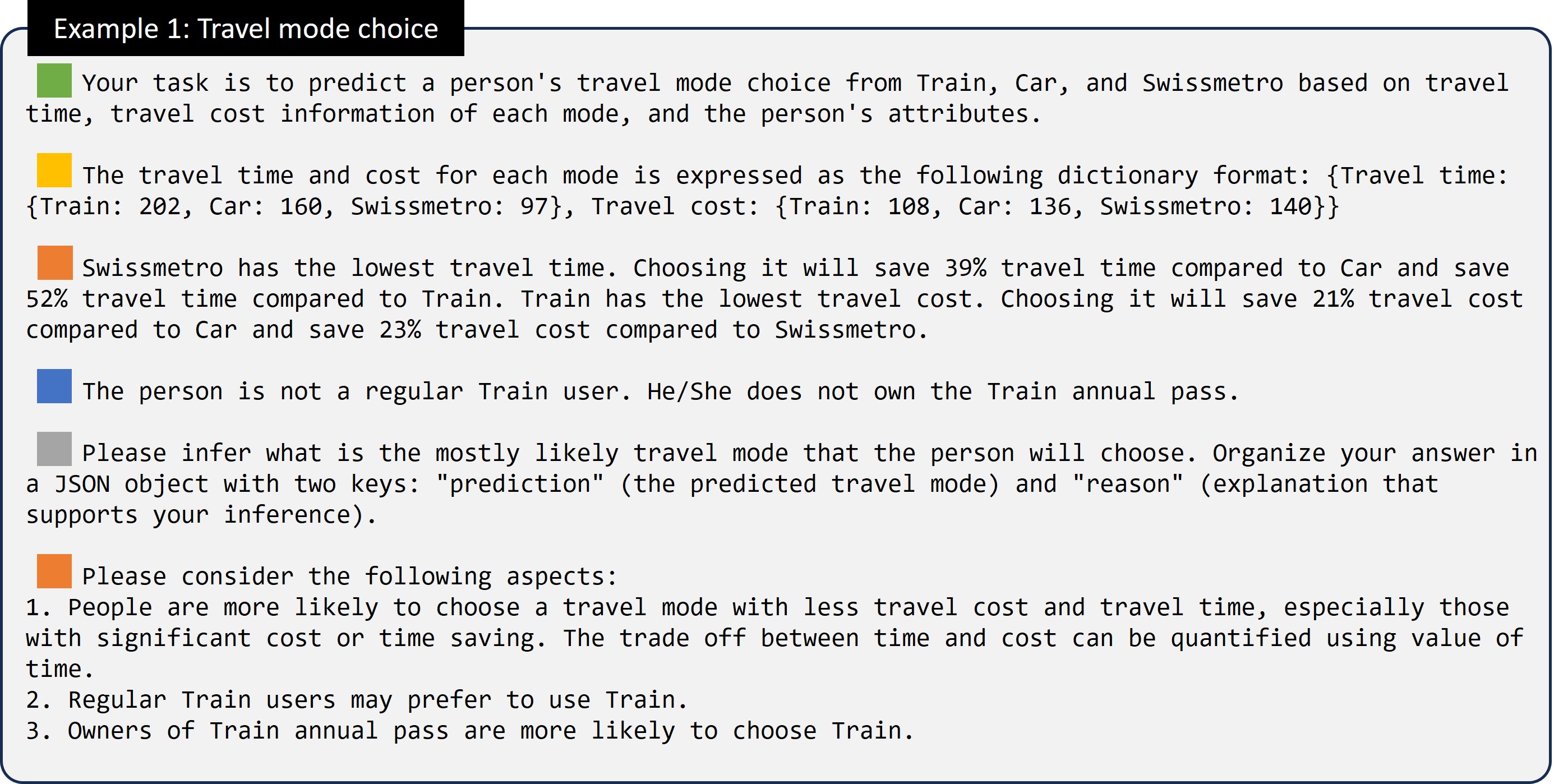}
    \caption{Example complete prompts for travel behavior prediction (Colors before each sentence represent different components. They will not be input in the LLMs)}
    \label{fig_example_sum}
\end{figure}

An example of the final prompt is shown in Figure \ref{fig_example_sum}. The sequence for different components is reorganized. It is worth noting that we do not include any training data information in the prompt (i.e., zero-shot), which allows the model to be used in any new context for travel behavior prediction. 

\subsection{Embedding as features}

Large Language Models (LLMs) encode semantic and contextual information into high-dimensional embedding vectors that can be leveraged for downstream prediction tasks. In the context of travel behavior prediction, we extract an embedding $\boldsymbol{h}_n$ for each sample $n$—typically representing a traveler, trip record, or textual description of the travel context. This embedding $\boldsymbol{h}_n$ serves as a compact yet expressive representation of the traveler's behavioral features, preferences, and context.

To predict the travel behavior label $y_n$ (e.g., mode choice, trip purpose, or destination), we use $\boldsymbol{h}_n$ as input to a supervised learning model parameterized by $\boldsymbol{\beta}$. The model is trained to minimize the empirical loss over the training dataset $\mathcal{D}^{\text{Train}}$:

\begin{align} \boldsymbol{\beta}^* = \arg \min_{\boldsymbol{\beta}}\sum_{n\in\mathcal{D}^{\text{Train}}}\ell_n\left(\text{SupervisedModel}(\boldsymbol{h}_n;\boldsymbol{\beta}), y_n\right) \end{align}

Here, $\ell_n(\cdot,\cdot)$ denotes a sample-wise loss function that quantifies the discrepancy between the predicted and true labels. For classification tasks, $\ell_n$ is typically the cross-entropy loss, while for regression tasks, it may be the mean squared error or other appropriate metrics.

This approach decouples representation learning (handled by the LLM) from the prediction model, allowing flexibility in choosing downstream models—such as logistic regression, random forests, gradient boosting machines, or neural networks—based on task-specific requirements and computational considerations. Notably, the use of $\boldsymbol{h}_n$ as input enables generalization across diverse travel contexts, especially when the embeddings capture high-level behavioral semantics informed by large-scale pretraining.

\section{Case study}\label{sec_case}
\subsection{Data}
Mode choice prediction is evaluated on the Swissmetro stated preference survey data set \citep{bierlaire2001acceptance}. The survey aims to analyze the impact of travel modal innovation in transportation, represented by the Swissmetro, a revolutionary maglev underground system, against the usual transport modes represented by car and train. The data contains 1,004 individuals with 9,036 responses. Users are asked to select from three travel modes (train, car, and Swissmetro) given the corresponding travel attributes. Trip purpose prediction is evaluated using the 2017 US National Household Travel Survey (NHTS), which tracks daily trip, socio-demographic features, and trip purposes. A total of 4 trip purposes are included: working, social, recreation, and others. 

For both datasets, we use balanced sampling to randomly select 1,000 entries for the large training set, 10 for the small training set (training sets are for benchmark models only), and 200 for the testing dataset. Balanced sampling means we ensure the proportion of all labels (i.e., mode choice and trip purposes) are the same. Results are derived from the average of 5 random samples.

The features and data statistics (combined all training and testing datasets) are shown in Tables \ref{tab_stat} and \ref{tab_stat2}.
\begin{table}[H]
\centering
\caption{Swissmetro training and testing data statistics}\label{tab_stat}
{
\begin{tabular}{@{}lcccc@{}}
\toprule
\textbf{Features}                        & \textbf{Mean} & \textbf{Std} & \textbf{Min} & \textbf{Max} \\ \midrule
Train travel time (min)              &   161.2   & 77.3  &  31   &  1049 \\
Train cost  (Swiss franc)            &  87.4   &  65.1   & 8   &  576 \\
Car travel time (min)                 & 138.7    &  96.8   &  32  & 1560  \\
Car cost (Swiss franc)                       &  86.9  &  46.9 &  8   &  520 \\
Swissmetro travel time (min)         & 85.7  &  53.5  &  8   & 790    \\
Swissmetro cost  (Swiss franc)        &  108.3   & 82.7   &  11   &  768 \\
Regular train user (Yes = 1)    &   0.35   &   0.48  &  0   &  1   \\
Own train annual pass (Yes = 1) &   0.13   &  0.34   &   0  & 1    \\ 
\bottomrule
\end{tabular}
}
\end{table}

\begin{table}[H]
\centering
\caption{NHTS training and testing data statistics}\label{tab_stat2}
{
\begin{tabular}{@{}lcccc@{}}
\toprule
\textbf{Features}                        & \textbf{Mean} & \textbf{Std} & \textbf{Min} & \textbf{Max} \\ \midrule
With child (Yes = 1)              &   0.39   & 0.49  &  0   &  1 \\
Primary activity last week is work (Yes = 1)           &  0.89   &  0.31   & 0   &  1 \\
Trip duration less than 10 min (Yes = 1)                 & 0.39    &  0.48   &  0  & 1  \\
Trip duration greater than 25 min (Yes = 1)                 & 0.28    &  0.45   &  0  & 1  \\
Travel mode is walk/bike (Yes = 1)                       &  0.08  &  0.27 &  0   & 1 \\
Travel mode is drive (Yes = 1)                       &  0.90 &  0.31 &  0  &  1 \\
Travel mode is taxi/rideshare (Yes = 1)                       &  0.00  & 0.07 &  0   &  1 \\
Travel mode is transit (Yes = 1)                       &  0.01  &  0.12 &  0  &  1 \\
Travel on weekends (Yes = 1)  & 0.24  &  0.43  &  0  & 1    \\
Only one person in trip (Yes = 1)       &  0.59   & 0.49   &  0   &  1 \\ \bottomrule
\end{tabular}
}
\end{table}

\subsection{Baseline models}
The LLM-based prediction model is compared against three benchmark models: 1) multinomial logit (MNL), 2) random forest (RF), and 3) neural networks (NNs). MNL is the canonical travel behavior prediction tool, while random forest and NNs are proven to perform well empirically \citep{wang2024comparing}. Since the sample sizes are small, we do not include more advanced deep-learning models as benchmarks.

\subsection{Parameter settings}
The specific LLM employed in the experiment is GPT-3.5 (version gpt-3.5-turbo-1106), GPT-4 (version gpt-4-turbo-2024-04-09), Llama 3.1 8B, and Llama 3.1 70B, which are all widely used LLMs with open APIs. The final prompts used for the two case studies are attached in Appendix \ref{append_prompt}. We set the temperature to 0 to avoid randomness in the output. The benchmark models are implemented using the python sklearn package with the default hyper-parameters. 

\subsection{Results}
\subsubsection{Prediction performance}
All models are evaluated based on prediction accuracy and weighted F1-score. Results are shown in Table \ref{tab_res_metrics}. Training pipeline large and small represent supervised learning with 1000 and 10 as training sample size, respectively. While zero-shot for LLM means directly prompting without any training. We see that supervised learning models (such as MNL, RF, and NN) trained on large datasets achieve the best overall performance for both mode choice and trip purpose prediction, confirming the effectiveness of data-driven training when sufficient labeled data are available. However, their performance deteriorates substantially when the training dataset is small, with notable drops in both accuracy and weighted F1-score across all model types.

\begin{table}[H]
\small
\centering
\caption{Combined model prediction results}
\label{tab_res_metrics}
\begin{tabular}{l|l||cc|cc}
\hline
\multirow{2}{*}{\textbf{Model}} & \multirow{2}{*}{\textbf{Training pipeline}} & \multicolumn{2}{c|}{\textbf{Mode choice}} & \multicolumn{2}{c}{\textbf{Trip purpose}} \\
\cline{3-6}
& & Accuracy & F1 score & Accuracy & F1 Score \\
\hline
\hline
MNL & large & 0.605 ± 0.036 & 0.595 ± 0.034 & 0.418 ± 0.043 & 0.401 ± 0.043 \\
RF & large & 0.612 ± 0.031 & 0.613 ± 0.026 & 0.459 ± 0.046 & 0.451 ± 0.047 \\
NN & large &  0.635 ± 0.031 &  0.637 ± 0.028 &  0.463 ± 0.047 &  0.455 ± 0.049 \\
\hline
MNL & small &  0.469 ± 0.075 &  0.457 ± 0.080 & 0.313 ± 0.048 & 0.294 ± 0.058 \\
MNL + Embedding & small &  0.494 ± 0.061 &  0.486 ± 0.057 & 0.327 ± 0.025 & 0.299 ± 0.029 \\
RF & small & 0.455 ± 0.068 & 0.441 ± 0.074 &  0.250 ± 0.000 &  0.100 ± 0.000 \\ 
RF + Embedding & small & 0.509 ± 0.077 & 0.498 ± 0.081 &  0.311 ± 0.025 &  0.278 ± 0.039 \\ 
NN & small & 0.462 ± 0.044 & 0.455 ± 0.046 & 0.268 ± 0.042 & 0.213 ± 0.023 \\ 
NN + Embedding & small & 0.532 ± 0.038 & 0.529 ± 0.038 & 0.296 ± 0.027 & 0.275 ± 0.031 \\ 
\hline
LLM-GPT-3.5 & zero-shot &  0.586 ± 0.052 &  0.572 ± 0.058 &  0.402 ± 0.030 & 0.359 ± 0.040 \\
LLM-GPT-4 & zero-shot &  0.570 ± 0.054 &  0.569 ± 0.054 &  0.401 ± 0.014 & 0.381 ± 0.020 \\
LLM-Llama-3.1 8B & zero-shot &  0.551 ± 0.047 &  0.539 ± 0.041 &  0.322 ± 0.008 &  0.243 ± 0.011 \\
LLM-Llama-3.1 70B & zero-shot & 0.540 ± 0.038 &  0.532 ± 0.041 &  0.408 ± 0.022 &  0.343 ± 0.022 \\
\hline
\end{tabular}
\end{table}

In contrast, LLM-based zero-shot predictions—without using any training samples—can outperform supervised models trained on limited data. This highlights the practical value of LLMs in real-world scenarios where labeled data are scarce or unavailable. Despite the absence of task-specific training, LLMs are able to leverage their pretrained knowledge to deliver competitive predictions.

Furthermore, when small training datasets are available, training supervised models with LLM-generated text embeddings consistently improves performance compared to using manually engineered features alone. This suggests that LLM embeddings capture richer semantic information that is difficult to encode through traditional feature design, thereby enhancing model generalization under data-constrained settings. Overall, these findings indicate that LLMs and their embeddings offer a promising alternative or complement to conventional supervised learning, particularly in low-data regimes.

\subsubsection{Interpretability}

\begin{table}[H]
\small
\centering
\caption{Example cases of using LLMs for travel mode choice prediction}\label{tab_res_reason}
\resizebox{1.0\linewidth}{!}
{
\begin{tabular}{p{0.1\linewidth}p{0.9\linewidth}}
\toprule
\multicolumn{2}{l}{\textbf{Case 1}: \emph{Correct prediction and reasonable explanations}} \\ \midrule
\textbf{Input}:      &  \{Travel time: \{Train: 106, Car: 90, Swissmetro: 34\}, Travel cost: \{Train: 72, Car: 70, Swissmetro: 78\}\}. The person is not a regular Train user. He/She does not own the Train annual pass.     \\
\textbf{Prediction}: &  Swissmetro (MNL prediction is Car)         \\                       \textbf{Actual}: &  Swissmetro       \\
\textbf{Reason}:     & Based on the given information, the person is not a regular Train user and does not own the Train annual pass. Swissmetro offers the lowest travel time, saving 62\% compared to Car and 68\% compared to Train. Although Car has the lowest travel cost, the significant time saving of Swissmetro makes it the most likely choice for the person. \\ \midrule\midrule
\multicolumn{2}{l}{\textbf{Case 2}: \emph{Wrong prediction but reasonable explanations}}                                                                                                                   \\\midrule
\textbf{Input}:     &  \{Travel time: \{Train: 114, Car: 96, Swissmetro: 55\}, Travel cost: \{Train: 82, Car: 80, Swissmetro: 105\}\}. The person is not a regular Train user. He/She does not own the Train annual pass.          \\
\textbf{Prediction}: &  Car (MNL prediction is Car)         \\                       \textbf{Actual}: &  Train       \\                                                     
\textbf{Reason}:   &  Based on the given information, the person is not a regular Train user and does not own the Train annual pass. Considering the trade-off between travel time and cost, the person is likely to choose the Car as it has the lowest travel cost and the difference in travel time compared to Swissmetro is not significant enough to outweigh the cost savings.     \\\midrule\midrule
\multicolumn{2}{l}{\textbf{Case 3}: \emph{Wrong prediction and unreasonable explanations}   } \\\midrule
\textbf{Input}:     &  \{Travel time: \{Train: 124, Car: 96, Swissmetro: 59\}, Travel cost: \{Train: 83, Car: 80, Swissmetro: 133\}\}. The person is a regular Train user. He/She does not own the Train annual pass.  \\
 \textbf{Prediction}: &  Train (MNL prediction is Train)         \\    
\textbf{Actual}: &  Swissmetro       \\             
\textbf{Reason}:  &     The person is a regular Train user, and regular Train users prefer to use Train. Although Swissmetro has the lowest travel time, the person's preference for Train and {\color{red}the absence of a Train annual pass make it more likely for the person to choose Train over Swissmetro}. Additionally, Car has the lowest travel cost, but the person's regular use of Train and preference for it make it more likely for the person to choose Train over Car despite the cost difference.                                                 \\\midrule\midrule
\multicolumn{2}{l}{\textbf{Case 4}: \emph{Wrong prediction and hallucinated explanations}   } \\\midrule
\textbf{Input}:     &  \{Travel time: \{Train: 95, Car: 130, Swissmetro: 92\}, Travel cost: \{Train: 29, Car: 44, Swissmetro: 32\}\}. The person is not a regular Train user. He/She does not own the Train annual pass.       \\
 \textbf{Prediction}: &  Swissmetro (MNL prediction is Train)         \\    
\textbf{Actual}: &  Train    \\             
\textbf{Reason}:  &  Based on the given information, the person is not a regular Train user and does not own the Train annual pass. Swissmetro has the lowest travel time, saving 29\% compared to Car and 3\% compared to Train. Although Train has the lowest travel cost, {\color{red}the person's attributes suggest a preference for time saving}, making Swissmetro the most likely choice.\\
\bottomrule
\end{tabular}
}
\end{table}
A unique advantage of LLM-based prediction is the interpretability of the results. Table \ref{tab_res_reason} shows several example outputs from the LLM. In Case 1, LLM can correctly predict the travel mode choice (while MNL cannot) and provide reasonable explanations. In Case 2, though LLM makes the wrong prediction, the explanations are reasonable. It talks about the reason for choosing a Car due to its lowest travel cost, and its travel time is not too different from other modes. In Case 3, we observe LLM's reasoning errors, stating that the absence of a train annual pass makes the person choose the train over Swissmetro. In Case 4, we see a common hallucination problem of LLMs \citep{Bang_Cahyawijaya_2023}. It claims that the person prefers time-saving over cost-saving without any related inputs. 

\subsubsection{Ablation studies}
In this section, we show how the exclusion of different components in the prompt would affect the performance. For simplicity, only the ablation of mode choice cases is tested with two GPT models. Note that, the ablation study still keeps all the information (same as the information given to supervised learning model), but with different prompt design. The meaning of different test cases are as follows: 
\begin{itemize}
    \item \textbf{Exclude guide of thinking}: We remove the sentences of ``Swissmetro has the lowest travel time. Choosing it will save 39\% travel time compared to ...'' from the prompt, which is used to guide the LLM to compare cost and travel time.
    \item \textbf{Exclude domain knowledge}: We remove the sentences of ``Please consider the following aspects:  1. People are more likely to choose ...'' from the prompt, which provides domain knowledge on how people typically make their decisions.
    \item \textbf{Change structural data to non-structural data}: Instead of using structural data as inputs (for example, \{Travel time: \{Train: 202, Car: 160, Swissmetro: 97\}\}), we directly tell the LLM about this information with descriptive sentences (e.g., the travel time of train is 202 minutes).
\end{itemize}

\begin{table}[H]
\centering
\caption{Ablation study results for the Swissmetro model choice dataset}
\label{tab_ablation}
\resizebox{1.0\linewidth}{!}
{
\begin{tabular}{l l cc}
\toprule
\multirow{2}{*}{\textbf{Model}} & \multirow{2}{*}{\textbf{Prompt design}} & \multicolumn{2}{c}{{\textbf{Performance}}} \\
\cmidrule(lr){3-4}
 &  & {Accuracy (Decrease \%)$^{1}$} & {F1 Score} \\
\midrule
\multirow{4}{*}{LLM-GPT-3.5} & Full prompt                & 0.586 ± 0.052 & 0.572 ± 0.058 \\
 & Exclude guide of thinking           & 0.552 ± 0.036 (-5.8\%) & 0.526 ± 0.039 \\
 & Exclude domain knowledge           & 0.583 ± 0.052 (-0.5\%) & 0.571 ± 0.052 \\
 & Change structural data to non-structural data   & 0.560 ± 0.037 (-4.4\%) & 0.538 ± 0.040 \\
\midrule
\multirow{4}{*}{LLM-GPT-4}  & Full prompt                 & 0.570 ± 0.054 &  0.569 ± 0.054\\
  &  Exclude guide of thinking          & 0.565 ± 0.044 (-0.8\%) & 0.562 ± 0.043\\
   & Exclude domain knowledge           & 0.569 ± 0.050 (-0.2\%) & 0.568 ± 0.052\\
   & Change structural data to non-structural data            & 0.567 ± 0.044 (-0.5\%) & 0.565 ± 0.043\\
\bottomrule
\multicolumn{4}{l}{
\begin{tabular}[c]{p{0.97\linewidth}} 
$^1$: Values in the parenthesis represents the mean accuracy decrease compared to the full prompt. 
\end{tabular}
} 
\end{tabular}
}
\end{table}

As shown in Table \ref{tab_ablation}, the ablation results indicate that prompt design plays a markedly different role for LLM-GPT-3.5 and LLM-GPT-4. For LLM-GPT-3.5, excluding the guide of thinking leads to the largest performance degradation, with accuracy dropping by 5.8\%, suggesting that explicit comparative reasoning cues are critical for this model to effectively weigh attributes such as travel time and cost. Changing structural data to non-structural, descriptive inputs also results in a performance decrease (4.4\%), highlighting the importance of well-organized, structured information for LLM-GPT-3.5. In contrast, excluding domain knowledge has a negligible impact (0.5\% decrease), which may indicate that such decision-making heuristics are already embedded in the model’s general reasoning capability. For LLM-GPT-4, performance remains largely stable across all ablation settings, with accuracy decreases below 1\% in all cases. This robustness suggests that GPT-4 is less sensitive to prompt engineering details and can effectively extract and utilize relevant information regardless of explicit reasoning guidance, domain knowledge cues, or data representation format, reflecting its more advanced reasoning and information integration capabilities.

\section{Conclusion and future work}\label{sec_conclusion}
This study investigates the potential of large language models (LLMs) as an alternative and complement to conventional travel behavior prediction methods. We propose and evaluate two LLM-based frameworks: a zero-shot prompting approach that directly generates predictions without task-specific training data, and a hybrid approach that integrates LLM-generated text embeddings with classical supervised learning models. Empirical results on mode choice and trip purpose prediction demonstrate that while traditional supervised models such as MNL, random forest, and neural networks achieve the best performance when large labeled datasets are available, their effectiveness degrades substantially in low-data regimes. In contrast, LLM-based zero-shot predictions achieve competitive performance without any training data and consistently outperform supervised models trained on very small samples. Moreover, incorporating LLM embeddings into supervised models significantly improves prediction accuracy and F1 scores under data-scarce conditions, indicating that LLMs capture high-level semantic information that is difficult to encode through manual feature engineering.

Beyond predictive performance, LLMs offer a distinct advantage in interpretability by providing natural language explanations for their predictions. Our qualitative analysis shows that these explanations are often reasonable and aligned with travel behavior theory, even when the predictions are incorrect. However, we also observe cases of reasoning errors and hallucinations, underscoring the need for caution when using LLM-generated explanations in decision-critical applications. The ablation study further highlights the importance of prompt design, particularly for less advanced models such as GPT-3.5, where explicit reasoning guidance and structured inputs play a significant role in prediction quality. In contrast, GPT-4 exhibits strong robustness to prompt variations, suggesting improved internal reasoning and information integration capabilities.

This work represents an initial exploration of LLMs for travel behavior prediction, and several directions remain for future research. First, more extensive experiments are needed using larger and more diverse datasets, including repeated trials to reduce randomness and evaluations across additional travel behavior tasks, such as departure time choice, route choice, and activity scheduling. Second, while this study includes an ablation analysis of prompt components, more systematic investigations into prompt optimization and standardization are needed to better understand how different forms of reasoning guidance, domain knowledge, and data representation affect LLM performance. Third, future work can explore few-shot and in-context learning strategies, where carefully selected representative samples are provided to guide LLMs in learning quantitative trade-offs such as the value of travel time or cost sensitivity from limited data. Finally, broader comparisons across a wider range of open-source and proprietary LLMs, as well as investigations into methods for reducing hallucinations and improving explanation faithfulness, would further clarify the practical applicability of LLMs in transportation modeling.

\appendix
\appendixpage
\section{Prompts}\label{append_prompt}
The final prompts for the travel mode prediction and trip purpose predictions are shown in Table \ref{tab_prompt}. Both are presented with an example. All changeable parts are underscored,  where different samples will have different values. 

\begin{table}[H]
\centering
\caption{Final prompts used for predictions}\label{tab_prompt}
{
\begin{tabular}{p{1.0\linewidth}}
\toprule
{\emph{\textbf{Prompt for travel mode prediction}}} \\ \hline
{Your task is to predict a person's travel mode choice from Train, Car, and Swissmetro based on travel time, travel cost information of each mode, and the person's attributes.

The travel time and cost for each mode is expressed as the following dictionary format: 
\{Travel time: \{Train: \underline{94}, Car: \underline{56}, Swissmetro: \underline{36}\}, Travel cost: \{Train: \underline{62}, Car: \underline{36}, Swissmetro: \underline{71}\}\}

\underline{Swissmetro} has the lowest travel time, Choosing it will save \underline{36\%} travel time compared to \underline{Car} and save \underline{62\%} travel time compared to \underline{Train}. 
\underline{Car} has the lowest travel cost. Choosing it will save \underline{49\%} travel cost compared to \underline{Swissmetro} and save \underline{42\%} travel cost compared to \underline{Train}.

The person \underline{is} a regular Train user. He/She \underline{does not own} the Train annual pass.

Please infer what is the mostly likely travel mode that the person will choose. Organize your answer in a JSON object with two keys: "prediction" (the predicted travel mode) and "reason" (explanation that supports your inference).

Please consider the following aspects: 

1. People are more likely to choose a travel mode with less travel cost and travel time, especially those with significant cost or time saving. The trade off between time and cost can be quantified using value of time. 

2. Regular Train users prefer to use Train. 

3. Owners of Train annual pass are more likely to choose Train.} \\ \midrule

{\emph{\textbf{Prompt for trip purpose prediction}}} \\ \hline
{Your task is to infer the purpose of a person's trip, classifying it into one of four categories: ``working'', ``shopping'', ``social'', and ``others''. You will make the inference based on the following information related to the socio-demographic information and trip information of the traveler.

The person lives in a household \underline{with} child. As of last week, His/her primary activity \underline{is} working. 

The trip happens on \underline{weekdays} and starts \underline{after 5pm}. The trip lasts \underline{less than 10 minutes}. The mode of transportation is \underline{not walking or biking}. There were \underline{2 or more people} on the trip.

Please also consider the following reasoning logics to support your inference:

1. Trips that happen on weekends are more likely to be social or shopping trips

2. Trips that happen before 10am are more likely to be working trips. Trips that happen after 5pm are more likely to be social trips

3. A trip with only one person is highly likely to be a working trip

4. Trips with walking or biking transportation mode are highly likely to be social trips

5. Working and social trip duration is usually longer than 25 min

6. People with child are more likely to trip purpose of "others", such as school or daycare

7. If the person's primary activity last week is working, it is more likely to be a working trip

Please follow the above logics strictly, especially not violate the rules. Infer what is the mostly likely trip purpose. Organize your final answer in a JSON object with two keys: "prediction" (the predicted travel purpose) and "reason" (explanation that supports your inference in one paragraph).
} \\ 
\bottomrule
\end{tabular}
}
\end{table}

\bibliography{mybibfile}

\end{document}